\documentclass[12pt]{article}
\usepackage[final,nonatbib]{neurips}

\usepackage[dvipsnames]{xcolor}
\usepackage{graphicx}
\usepackage{subcaption}
\usepackage{pifont}
\usepackage{amsmath,amssymb}
\usepackage{algorithm} 
\usepackage{algpseudocode} 
\usepackage{centernot}

\usepackage{tikz}
\usepackage{comment}
\usepackage{booktabs}
\usepackage{color}
\usepackage{etoolbox}
\usepackage[normalem]{ulem}

\usepackage[colorlinks=true]{hyperref}
\usepackage{cleveref}

\begin{document}
\pagestyle{headings}
\title{How stable are Transferability Metrics evaluations?}

\author{
Andrea Agostinelli \And
Michal Pándy\AND
Jasper Uijlings \And
Thomas Mensink \And
Vittorio Ferrari\\
}
\date{Google Research\thanks{Contact: \texttt{\{agostinelli,michalpandy,jrru,mensink,vittoferrari\}@google.com}}}

\maketitle

\providetoggle{showcomments}
\settoggle{showcomments}{true} 								%

\iftoggle{showcomments}{%
    \newcommand{\changed}[1]{\textcolor{blue}{#1}}
    \newcommand{\todo}[1]{\textcolor{blue}{\textbf{TODO:} #1}}
    \newcommand{\resolved}[3][]{\ifstrequal{#1}{resolved}{\textcolor{blue}{RESOLVED:}~\textbf{{\MakeUppercase #2:}}~{#3}}{\textbf{\MakeUppercase #2:}~#3}}
    \newcommand{\andrea}[2][]{\textcolor{ForestGreen}{\resolved[#1]{andrea}{#2}}}
    \newcommand{\vitto}[2][]{\textcolor{red}{\resolved[#1]{vitto}{#2}}}
    \newcommand{\jasper}[2][]{\textcolor{violet}{\resolved[#1]{jasper}{#2}}}
    \newcommand{\michal}[2][]{\textcolor{pink}{\resolved[#1]{michal}{#2}}}
    \newcommand{\tm}[2][]{\textcolor{magenta}{\resolved[#1]{TM}{#2}}}
    \newcommand{\att}[1]{\textcolor{red}{#1}}
}{%
    \newcommand{\changed}[1]{#1}
    \newcommand{\todo}[1]{}
    \newcommand{\andrea}[2][]{}
    \newcommand{\vitto}[2][]{}
    \newcommand{\jasper}[2][]{}
    \newcommand{\tm}[2][]{}
    \newcommand{\att}[1]{#1}
}
\newcommand{\TM}[2][]{\tm[#1]{#2}}
\newcommand{\DONE}{\done}
\newcommand{\leep}{LEEP}
\newcommand{\nleep}{$\mathcal{N}$LEEP}
\newcommand{\logme}{LogME}
\newcommand{\gbc}{GBC}
\newcommand{\hscore}{H-score}
\newcommand{\para}[1]{
    \par\noindent\textbf{#1}
}

\definecolor{ForestGreen}{RGB}{34,139,34}

\newcommand{\cmark}{\ding{51}}%
\newcommand{\xmark}{\ding{55}}%

\DeclareRobustCommand\legendsource{\tikz[baseline=-.04mm] \draw[NavyBlue, very thick] (0,0.1) -- (0.3,0.1);}
\DeclareRobustCommand\legendtarget{\tikz[baseline=-.03mm] \draw[ForestGreen, very thick] (0,0) -- (0,0.3);}
\DeclareRobustCommand\legendevaluation{\tikz \draw[BrickRed, very thick] (0.1,0.1) -- (0.3,0.3);}

\begin{abstract}
    Transferability metrics is a maturing field with increasing interest, which aims at providing heuristics for selecting the most suitable source models to transfer to a given target dataset, without fine-tuning them all.
    However, existing works rely on custom experimental setups which differ across papers, leading to inconsistent conclusions about which transferability metrics work best.
    In this paper we conduct a large-scale study by systematically constructing a broad range of 715k experimental setup variations. We discover that even small variations to an experimental setup lead to different conclusions about the superiority of a transferability metric over another.
    Then we propose better evaluations by aggregating across many experiments, enabling to reach more stable conclusions. 
    As a result, we reveal the superiority of \logme{} at selecting good source datasets to transfer from in a semantic segmentation scenario,  \nleep{} at selecting good source architectures in an image classification scenario, and \gbc{} at determining which target task benefits most from a given source model. Yet, no single transferability metric works best in all scenarios.
\end{abstract}
\section{Introduction}

Transfer learning aims to re-use knowledge learned on a source task to help learning a target task, for which typically there is only little training data.
The most prevalent method of transfer learning in computer vision is to pre-train a source model on a large source dataset (\emph{e.g.}, ILSVRC'12~\cite{russakovsky15ijcv}), and then fine-tune it on the target dataset \cite{azizpour15pami,chu16eccv,girshick15iccv,he17iccv,huh16nips,shelhamer16pami,zhou19arxiv}.
However, different target tasks benefit from using different source model architectures~\cite{chen18pami,he16cvpr,newell16eccv,ronneberger15miccai} or pre-training on different source datasets~\cite{mensink21arxiv,ngiam18arxiv,yan20cvpr}.
Hence, a key challenge is determining which source model is best suited for which target task, and doing so in a computationally efficient manner. 

Transferability metrics~\cite{agostinelli22ensembles,bao19icip,bolya21neurips,dwivedi20eccv,dwivedi19cvpr,li21cvpr,nguyen20icml,pandy2022transferability,song19neurips,song20cvpr,tan21cvpr,tran19iccv,you21icml}
provide heuristics for selecting the most suitable source models for a given target dataset, without explicitly fine-tuning them all. These methods generally work by applying a source model to the target dataset to compute embeddings or predictions. Then they efficiently assess how compatible these embeddings/predictions are with the target labels. This provides a proxy for how well the source model transfers to the target task.

Transferability metrics is a maturing field with numerous contributions and increasing interest~\cite{agostinelli22ensembles,bao19icip,li21cvpr,nguyen20icml,pandy2022transferability,tan21cvpr,tran19iccv,you21icml}. However, existing works rely on custom experimental settings without standardized benchmarks, which leads to inconsistent conclusions across different papers.
In particular, contradicting conclusions often appear when comparing findings across papers.
For example, while NCE \cite{tran19iccv} consistently outperforms LEEP \cite{nguyen20icml} in \cite{you21icml}, LEEP outperforms NCE in both \cite{tan21cvpr} and \cite{nguyen20icml}.
Furthermore \logme{} outperforms LEEP \cite{you21icml}, but LEEP outperforms \logme{} in \cite{pandy2022transferability}.
This raises questions about how stable conclusions are across different experimental setups and if there truly is a single best transferability metric.

The primary goal of our work is to evaluate the stability of experimental setup for transferability estimation. We observe that a single experiment typically consists of three components: 
(1) a choice of the pool of source models;
(2) a choice of the target dataset;
(3) a choice of the measure used to evaluate how well transferability metrics perform.
We vary each of these components in a large-scale systematic study of two scenarios: selecting good source datasets for semantic segmentation
(Sec.~\ref{sec:semseg}), and selecting good source model architectures for image classification (Sec.~\ref{sec:architecture-selection}).
In total we construct a large set of \textbf{715k experiments}, several orders of magnitude larger than previous works~\cite{nguyen20icml,li21cvpr,you21icml,pandy2022transferability,bao19icip} (and in a computationally efficient way).
The source code for our experiments and analysis is publicly available\footnote{\href{https://github.com/google-research/google-research/tree/master/stable_transfer}{\texttt{github.com/google-research/google-research/tree/master/stable\_transfer}}}.
Based on these experiments: 
(A) We demonstrate that even small variations to an experimental setup leads to very different conclusions about the superiority of a transferability metric over another.
(B) We provide a systematic analysis to investigate which of the three setup components contributes the most to the instability of experimental outcomes.
(C) We propose better evaluations by aggregating outcomes from a broad set of diverse experiments, reducing the experimental uncertainty and enabling to reach more stable conclusions.
Concretely, we reveal that \logme{} is the best transferability metric in our first scenario (selecting source datasets for semantic segmentation); \nleep{} is the best in our second scenario (selecting source model architectures for image classification).
Moreover, we also consider a third, somewhat separate scenario about determining which target task benefits most from a given source model, among a pool of target tasks constructed by subsampling out of a large target dataset (Sec.~\ref{sec:target-selection}).
In this scenario \gbc{} is the best metric.
So, no single transferability metric works best in all scenarios.

\section{Related Work}

\para{Transfer learning.}
Training a deep neural network for a specific task often requires a large amount of data which can be difficult to obtain. The goal of transfer learning \cite{bozinovskiinfluence,pratt1992discriminability,thrun2012learning} is to leverage information from a source task with easily obtainable data, to improve performance on a target problem where data is scarce. Pre-training a neural network on large datasets and fine-tuning it on a target dataset is the most prevalent method of transfer learning in computer vision. For these reasons, there exists a wide array of large-scale source datasets, such as ILSVRC'12~\cite{russakovsky15ijcv}, ImageNet21k~\cite{ridnik2021imagenet}, or Open Images~\cite{kuznetsova20ijcv}. In addition, recent works consider the use of unlabeled source datasets via self-supervised pre-training \cite{chen20icml,chen2021exploring,he2020momentum}. 
Other research studies settings in which transfer learning is effective. Taskonomy \cite{zamir18cvpr} develops connections between visual tasks (\emph{e.g.} semantic segmentation, depth prediction, etc.), Mensink et al. \cite{mensink21arxiv} perform extensive experimental investigations in semantic segmentation settings, Mustafa et al. \cite{mustafa2021supervised} focus on evaluating transfer learning for medical purposes, and Ding et al. \cite{ding2019empirical} explore applications to human activity recognition. For a more general survey of transfer learning, we refer the reader to Weiss et al. \cite{weiss2016survey}.
\para{Transferability metrics.} 
Transferability metrics provide efficient heuristics for determining which pre-trained models are most suitable for a specific target task. To generate transferability metrics, \textit{label comparison-based methods} leverage the labels of the source and target domains. These methods generally assume equivalence between source and target domain labels or obtain pseudo-labels by executing the source model on the target domain. Methods in this category include NCE \cite{tran19iccv} and LEEP \cite{nguyen20icml}. \textit{Source embedding-based} methods use the feature extractor of a pre-trained neural network to embed target domain images. Transferability metrics are then computed using the embeddings and their corresponding labels. These methods include GBC \cite{pandy2022transferability}, LogME \cite{you21icml}, and \nleep{} \cite{li21cvpr}. To conclude, \textit{optimal transport-based} methods \cite{tan21cvpr,alvarez2020geometric} develop cost functions usable within the optimal transport framework to determine transferability between two datasets. See Section \ref{sec:background} for a more detailed description of state-of-the-art tranferability metrics.
\para{Overview of experimental stability.}
Several works study experimental settings and analyse methods' performance with the aim of solidifying an area of research. In the field of graph neural networks (GNNs), Errica et al. \cite{errica2019fair} evaluate and compare GNNs across a large suite of experiments, while Dwivedi et al. \cite{dwivedi2020benchmarking} and Shchur et al. \cite{shchur2018pitfalls} point out issues with GNN evaluation and propose evaluation improvements. In reinforcement learning (RL), Whiteson et al. \cite{whiteson2011introduction} discuss RL evaluation principles, Jordan et al. \cite{jordan2020evaluating} discover flaws with RL evaluation metrics, and Colas et al. \cite{colas2018many} study the necessary number of seeds for stable RL testing. In computer vision, Hoiem et al. \cite{hoiem2012diagnosing} and Hosang et al. \cite{hosang2015makes} analyse the stability of object detectors, Zendel et al. \cite{zendel2017good} discuss the quality of testing data, Xian et al. \cite{xian19pami_good_bad_ugly} evaluate state-of-the-art in zero-shot learning, and Abnar et al. \cite{abnar2021exploring} assess the trade-offs of upstream and downstream model performance. Our work provides clarity about good experimental settings for research in transferability metrics.

\section{Background}
\label{sec:background}

We first discuss the basics of transferability metrics and how they are evaluated. At a high level, the overall pipeline in most papers is~\cite{agostinelli22ensembles,bao19icip,li21cvpr,nguyen20icml,pandy2022transferability,tan21cvpr,tran19iccv,you21icml}:
(1)
a transferability metric considers a source model $S$ and a target dataset $T$ and predicts a transferability score $M$. $M$ predicts how well $S$ will transfer to $T$ (Sec.~\ref{sec:transferability_metrics}).
(2)
compute the true accuracy of transferring from $S$ on $T$ by fine-tuning it on the target training set, then applying the fine-tuned model on the target test set, and finally evaluate accuracy $A$ based on the ground-truth (Sec.~\ref{sec:model_transferability}).
(3)
evaluate the quality of the transferability metric by checking how well $M$ predicted true accuracy $A$ (Sec.~\ref{sec:eval_measures}).

\subsection{Transferability metrics}
\label{sec:transferability_metrics}

In our paper we compare the following transferability metrics M.

\para{\hscore{}.}
\hscore{}~\cite{bao19icip} is based on the intuition that a model transfers well to a target dataset if the target embeddings have low inter-class variance and low feature redundancy. These quantities are computed by constructing the inter-class and data covariance matrices.

\para{\leep{} \& \nleep{}.}
\leep{}~\cite{nguyen20icml} first predicts pseudo-labels by applying the source model on the target images. These predictions are then used for computing a log-likelihood between the target labels and the source model predictions. The core idea is that if the predictions are `clustered' around individual target labels, adaptation to the target dataset should be easier.
\nleep{}~\cite{li21cvpr} extends this idea by first embedding the target images using the source feature extractor and then fitting a Gaussian mixture model (GMM) on them. The GMM is said to provide a better density estimator for the pseudo-labels than the classification head in the original \leep{}.

\para{\logme{}.}
After embedding the target images using the source feature extractor, \logme{}~\cite{you21icml} computes the probability of the target labels conditioned on these embeddings (i.e. the evidence of target labels). By setting up a graphical model and using independence assumptions between samples, the authors propose an efficient algorithm for computing such evidence.

\para{\gbc{}.}
\gbc{}~\cite{pandy2022transferability} measures the statistical overlap between classes of the target dataset, after representing the target images in the embedding space determined by the source feature extractor. The intuition is that the more classes overlap in that space, the more difficult it will be to achieve high accuracy after fine-tuning the source model. The overlap is estimated using the Bhattacharyya coefficient between multivariate Gaussians fitted to each class.

\subsection{True accuracy of transfer learning}
\label{sec:model_transferability}

We compute the true accuracy $A$ of the source model $S$ on the target test set after fine-tuning on the target training set. $A$ represents how well $S$ transfers to the target $T$.
We note that we fine-tune the full source model on the target training dataset, rather than just the classification head.

\subsection{Evaluating the quality of a transferability metric}
\label{sec:eval_measures}

We introduce common measures for evaluating transferability metrics. Suppose we wish to evaluate a transferability metric $M$ given a source pool $\mathcal{S}$ containing $n$ source models, and a target dataset $T$.
Hence, we have access to $n$ transferability metric values $M_i$, each with its corresponding test accuracy $A_i$ associated with source model $S_i$.
An evaluation measure captures how well $M_i$ relates to $A_i$ (i.e. higher values of transferability $M_i$ predict higher values of true accuracy $A_i$).

\para{Pearson correlation coefficient \cite{benesty2009pearson} ($\rho$).}
It measures linear correlation between $M_i$ and $A_i$ across $i \in [1 \mathrel{{.}\,{.}}\nobreak n]$. For transferability metrics that tend to be linear with respect to test scores (e.g, LEEP \cite{nguyen20icml}), it provides a straightforward way for measuring performance. The disadvantage is that low $\rho$ correlation does not imply a bad a performing transferability metric. The transferability rankings could be correct even though $M_i$ and $A_i$ are non-linearly related.

\para{Kendall rank correlation coefficient \cite{kendall1938new}  ($\tau$).}
The core idea is that we should have $M_i > M_j$ if $A_i > A_j$. Based on this, $\tau$ is computed as:
\begin{equation}
    \label{eq:tau-def}
    \tau = \frac{1}{\binom{n}{2}} \sum_{i < j} \text{sgn}(M_i - M_j) * \text{sgn}(A_i - A_j)
\end{equation}
Eq.~\eqref{eq:tau-def} can be interpreted as averaging over the agreements ($+1$) or disagreements ($-1$) in rankings between transferability metrics and test scores. Hence, $\tau \in [-1, 1]$ and high $\tau$ implies a strong correlation between the rankings of source models according to $M_i$ and according to $A_i$.
$\tau$ does not rely on linearity assumptions and gives a holistic idea for the overall ranking performance.
The downside is that $\tau$ gives equal importance to all source models, regardless of their performance.
However, practitioners ultimately care about a transferability metric's ability to correctly rank only the few best performing models.

\para{Weighted Kendall rank correlation coefficient ($\tau_w$).}
To prioritize the top-performing models, we may assign higher weights to them. Models with higher $A_i$ within the sum in Eq.~\eqref{eq:tau-def} should be weighted more.
Hence, two transferability metrics that would perform the same in terms of $\tau$ can now be discriminated based on which metric ranks top models better in $\tau_w$.
We still have $\tau_w \in [-1, 1]$ and now the evaluation measure reflects practitioners' priorities. We assume a hyperbolic drop-off in model importance with decreasing ranks in terms of $A_i$, which is implemented in the Sklearn library \cite{scikit-learn}.

\para{Relative top-1 accuracy \cite{li21cvpr} ($Rel@1$).}
It measures how close model $k$ with the highest predicted transferability ($k=\text{argmax}_i \ M_i$) performs, in terms of accuracy $A_k$, compared to the highest performing model ($\max_i A_i$).
It is computed as $Rel@1 = \frac{A_k}{\max_i A_i}$.
Since $A_k \leq \max_i A_i$, we have $1 \geq Rel@1 \geq 0$. 
The benefit of $Rel@1$ is that there could be multiple models that perform within a close margin of the top model, which in terms of $Rel@1$ would obtain similar performance scores. Since $\tau$ and $\tau_w$ are not sensitive to the actual value of the test scores of models, the evaluation of a particular transferability metric could vary drastically different even if similar-performing source models change order in the ranked lists (which can happen even for small fluctuations in the values of $M_i$ or $A_i$).

\section{Methodology} 

\begin{figure}[t]
    \centering
    \includegraphics[trim=20 90 30 80, clip, width=\textwidth]{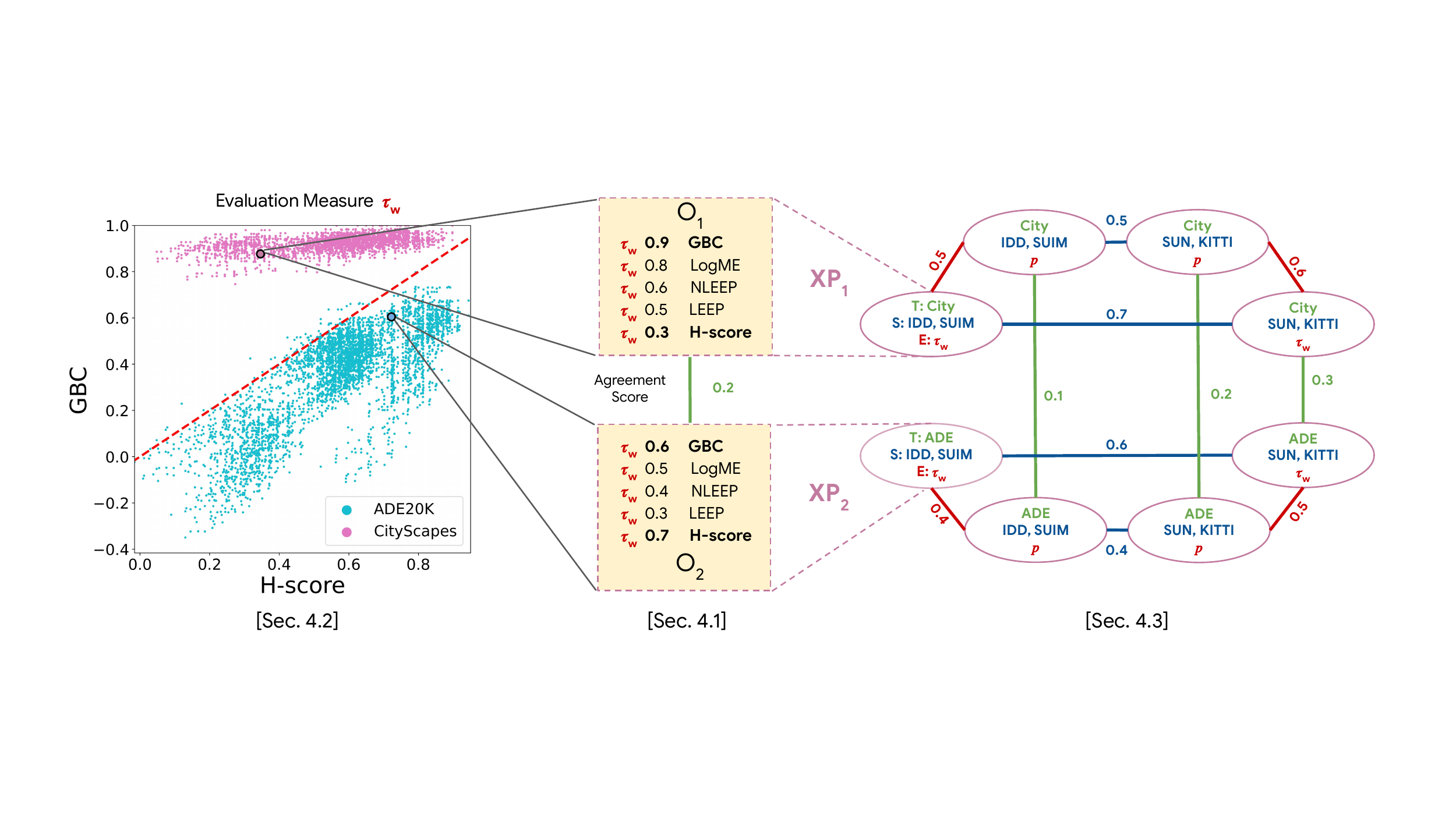}
    \caption{
    An experiment $\textrm{XP}$ is represented as a graph node, composed of source pool $\mathcal{S}$, target dataset $T$ and evaluation measure $E$. We compare two experiments $\textrm{XP}_1$ and $\textrm{XP}_2$ by comparing their outcomes $O_1$ and $O_2$.
    We do it qualitatively with a scatter plot (left) and quantitatively with a graph (right).
    For the latter we connect experiments that differ by a single variation in either $\mathcal{S}$ (\legendsource), $T$ (\,\legendtarget\,) or $E$ (\legendevaluation).
    We compute the agreement score between the outcomes of each two connected experiments $(\textrm{XP}_i, \textrm{XP}_j)$ (edge value).}
    \label{fig:graph-robustness-score}
\end{figure}

Our goal is to investigate the experimental stability of protocols to evaluate transferability metrics.
We first rigorously define what constitutes a single experiment and identify its three main components (Sec. \ref{sec:experiment-components}).
We then investigate the effects of small variations on each of these components both qualitatively (Sec. \ref{sec:qualitative-method}, Fig. \ref{fig:graph-robustness-score} left) and quantitatively (Sec.\ref{sec:quantitative-method}, Fig. \ref{fig:graph-robustness-score} right).
Finally, in Sec. \ref{sec:700k-experiments} we define how we create a huge set of experiments that allows us to explore the influence of varying each component on the evaluation of transferability metrics. At the same time, the sheer size of our experiments helps to reduce experimental uncertainty and improve their stability.

\subsection{A single experiment with its three components}\label{sec:experiment-components}

Given a set of transferability metrics, the outcome $O$ of an experiment $\textrm{XP}$ is a measure of quality for each transferability metric (Fig.~\ref{fig:graph-robustness-score}, middle). 
A single experiment consists of three components whose choice influences its outcome:
(1) A \emph{pool of source models} $\mathcal{S}$. It can be created by training the same model architecture on different source datasets~\cite{agostinelli22ensembles,pandy2022transferability}, or by training different model architectures on the same source dataset~\cite{nguyen20icml,you21icml}, or a combination of both~\cite{li21cvpr}.
(2) A \emph{target dataset} $T$, which cannot be any dataset in the source pool.
(3) An \emph{evaluation measure} $E$ to determine the quality of each transferability metric.
Hence we can represent one experiment as $\textrm{XP}(\mathcal{S}, T, E) \rightarrow O$ (Fig.~\ref{fig:graph-robustness-score}).

We can compare two experiments $\textrm{XP}_i$ and $\textrm{XP}_j$ by looking at differences in their outcomes $O_i$ and $O_j$. For example, the experiments $\textrm{XP}_1$ and $\textrm{XP}_2$ are composed of the same source pool $\mathcal{S}$ (i.e. IDD, SUIM), the same evaluation measure $E$ (i.e. $\tau_w$), but different target datasets $T$ (i.e. ADE20k vs CityScapes). These experiments produce divergent outcomes, as \gbc{}$>$\hscore{} on $\textrm{XP}_1$, whereas \hscore{}$>$\gbc{} on $\textrm{XP}_2$.

We use these definitions to investigate how varying the components of an experimental setup ($\mathcal{S}$, $T$, $E$) affects the outcome $O$ across experiments. We note that an experimental setup is stable if small variations of $\mathcal{S}$,  $T$, or $E$ produce similar outcomes (and unstable if small variations produces divergent outcomes).

\subsection{Qualitative analysis of experimental setup stability}\label{sec:qualitative-method}

We qualitatively compare a large number of experiments by displaying their outcomes in a scatter plot (Fig.~\ref{fig:graph-robustness-score}).
Each point represents a single experiment and compares the quality of a transferability metric (\emph{e.g.} \hscore{}) to the quality of another (\emph{e.g.} \gbc{}) as assessed by a fixed evaluation measure $E$ ($\tau_w$ in this example).
Each experiment is colored according to its target dataset $T$. The dotted red line represents $x=y$; for points on this line both transferability metrics are equally good. If points are either all above or below the line, this means that one transferability metric consistently outperforms the other, suggesting the experiment is stable. %

We can study how variations in the source pool $\mathcal{S}$ influence the outcome by looking at the distribution of points of the same color (i.e. on the same target dataset $T$).
In Fig.~\ref{fig:graph-robustness-score}, in all experiments where the target dataset is CityScapes (magenta), \gbc{} consistently outperforms \hscore{}. Hence this experiment is stable with respect to variations in the source pool.
To study the influence of varying the target dataset $T$, we can compare whole point clouds (in  different colors). In this case, for $T=\textrm{ADE20k}$ \hscore{} consistently outperforms \gbc{}, which contradicts with what found for $T=\textrm{CityScapes}$. Hence this experiment setup is unstable when varying the target dataset.
Finally, we can compare the influence of varying the evaluation measure $E$ by comparing two scatter plots which differ only in the choice of $E$ (\emph{e.g.} Fig.~\ref{fig:source_datasets_influence_evaluation_measures} vs Fig.~\ref{fig:source_arch_influence_evaluation_measures}).

\subsection{Quantitative measures of experimental setup stability}\label{sec:quantitative-method}

\para{Setup Stability (SS).}
We want to quantify the overall effect of varying exactly a single component $\mathcal{S}$, $T$ or $E$ of an experimental setup. To do so, we consider pairs of experiments which differ in only one component.
Each edge in Fig.~\ref{fig:graph-robustness-score} (right) connects two experiments and is colored based on which component is different between them. Experiments which differ in two or more components are not connected.

We now detail how we measure the Setup Stability for a single component (\emph{e.g.} $T$ in green).
First, for each pair of experiments which differ only in $T$, we calculate an agreement score between their outcomes: the rank correlation $\tau$ between the two lists of transferability metrics as ranked by their quality %
For example, the agreement score between $\textrm{XP}_1$ and $\textrm{XP}_2$ in our example is $0.2$ (middle of Fig.~\ref{fig:graph-robustness-score}). Then we average agreement scores over all experiment pairs which differ only in $T$ (i.e. green edges in Fig.~\ref{fig:graph-robustness-score}, right side).
We call this aggregated measure the \emph{Setup Stability} (SS).
It is 1 when all experiments produce the same outcome, and indicates a stable setup. 
In contrast, a score of 0 implies there is no agreement at all across the experiment outcomes (very unstable).
In our example in Fig.~\ref{fig:graph-robustness-score}, the SS for $T$ equals to $\tfrac{1}{4}\left(.2 + .1 + .2 + .3 \right) = 0.20$.

\para{Win rate.}
We introduce the \emph{win rate} to assess whether a particular transferability metric is the best across many experiments.
The win rate of a transferability metric is the percentage of experiments where it is the best metric (i.e. outperforms all other metrics in that experiment).
We report the win rate for each of the four evaluation measures $E$ individually, which provides another view of its influence.
We also consider the win rate across all experiments, as the final, most aggregated assessment of a transferability metric.

\subsection{Constructing 715k experiments}\label{sec:700k-experiments}
We need to construct a set of experiments which vary along each of the three components $\mathcal{S}, T$, and $E$. This set needs to be extremely large to properly investigate the influence of each component and to be able to reduce experimental uncertainty when deciding which transferability metrics work best.
In Sec.~\ref{sec:semseg} and~\ref{sec:architecture-selection} we will study in-depth two scenarios about selecting good source models for the tasks of semantic segmentation and image classification. 
In total we construct {\em 715k experiments}.
This is several orders of magnitude more than other works on transferability metrics, which all have at most 500 experiments when selecting source models (Tab.~\ref{tab:setup_comparison}).

\begin{table}[t]
\footnotesize	%
  \footnotesize
  \centering
  \resizebox{\textwidth}{!}{
  \begin{tabular}{l l | c c c c c c}
     & & Ours  & \leep{} & \nleep{} & \logme{} & \gbc{} & OTCE \\
     & &  & \cite{nguyen20icml}
     & \cite{li21cvpr} & \cite{you21icml} & \cite{pandy2022transferability} & \cite{tan21cvpr}\\
     \textit{Selecting} & \#sources & 17  & 0  & 16 & 0 & 17 & 4\\
     \textit{source datasets} (Sec.\ref{sec:semseg}) & \#targets & 17  & 0  & 4 & 0 & 17 & 5\\
     \hline
     \textit{Selecting} & \#sources & 19  & 9 & 13  & 24 & 9 & 0\\
     \textit{source architectures} (Sec.\ref{sec:architecture-selection}) & \#targets & 9  & 1 & 4 & 17 & 8 & 0\\
     \hline
     \textit{Evaluation measures} (Sec.\ref{sec:eval_measures}) & \#$E$ & 4 & 1 & 6 & 1 & 2 & 1\\
     \hline
      \textit{Constructing XPs} (Sec.\ref{sec:700k-experiments}) & \# experiments & 715k & 1 & 48 & 19 & 50 & 500\\ \\
  \end{tabular}
  } %
  \caption{Volume of experiments for evaluating transferability metrics in various papers. We systematically construct 715k experiments by varying the source pool $\mathcal{S}$, target dataset $T$ and evaluation measure $E$. In contrast, previous works have at most 500 experiments.
  All stats are about the scenario of selecting source models for a given target. 
  Some papers \cite{nguyen20icml,bao19icip,tan21cvpr,pandy2022transferability,li21cvpr} also report the converse case (determining which target task benefits most from a given source model, which we consider in Sec.\ref{sec:target-selection}
  ). Even when including it, no previous work reports more than 508 experiments \cite{tan21cvpr}.
  The OTCE paper \cite{tan21cvpr} constructs 500 experiments by randomly sampling 100 target tasks within each of the 5 target datasets used.
  We note that the \hscore{} paper \cite{bao19icip} only reports in this converse scenario (and only 1 experiment).}.
  \label{tab:setup_comparison}  
\end{table}
\section{Scenario 1: selecting a good source dataset in semantic segmentation}
\label{sec:semseg}

\para{Experimental setup.}
\label{sec:semseg-experimental-setup}
We consider the scenario of selecting good source datasets to transfer from on the task of semantic segmentation.
We use a total of 17 datasets from a wide range of image domains: Pascal Context~\cite{mottaghi14cvpr}, Pascal VOC~\cite{pascal-voc-2012}, ADE20K~\cite{zhou17cvpr}, COCO~\cite{caesar18cvpr,lin14eccv,kirillov19cvpr}, KITTI~\cite{alhaija18ijcv}, CamVid~\cite{brostow09prl}, CityScapes~\cite{cordts16cvpr}, IDD~\cite{varma19wacv}, BDD~\cite{yu20cvpr}, MVD~\cite{neuhold17cvpr}, ISPRS~\cite{rottensteiner14isprs}, iSAID~\cite{waqas2019isaid,xia18cvpr}, SUN RGB-D~\cite{song15cvpr}, ScanNet~\cite{dai17cvpr}, SUIM~\cite{islam2020suim}, vKITTI2~\cite{cabon20arxiv,gaidon16cvpr} and vGallery~\cite{weinzaepfel19cvpr}.
We use a fixed model architecture composed of a HRNetV2-W48 backbone~\cite{wang20pami} with a linear classifier on top, as this model achieves state-of-the-art performances on dense prediction tasks~\cite{lambert20cvpr,mensink21arxiv,wang20pami}.
We create a total of 17 source models by training HRNetV2-W48 on the full training set of each source dataset.

\para{Setup variations.}
Now we create many setup variations.
First, each dataset plays the role of target, in turn.
For each target the other 16 datasets are sources, leading to 16 source models.
To construct many different source pools $\mathcal{S}$, we consider all combinations of 11 out of 16 source models, resulting in $\binom{16}{11}$ source pools.
Finally, we consider four evaluation measures $E$.
Hence we have a total of $17 \times \binom{16}{11} \times 4 = 297$k experiments, spanning variations across source pool $\mathcal{S}$, target dataset $T$ and evaluation measure $E$.
Importantly, while we consider a very large space of setup variations, this only requires training 17 source models and fine-tuning each $16$ times, which is computationally efficient.

When fine-tuning on the target training set, we follow the low-shot setup of~\cite{mensink21arxiv}: we limit it to 150 images for each dataset (except COCO and ADE20k, on which we use 1000 images as they contain a large number of classes).
We follow~\cite{agostinelli22ensembles} and apply the transferability metrics to semantic segmentation by subsampling 1000 pixels per image. As in~\cite{agostinelli22ensembles}, we sample pixels inversely proportionally to the frequency of their class labels in the target dataset.

\begin{figure}[tp]
    \centering
    \includegraphics[trim=10 120 10 110, clip, width=\textwidth]{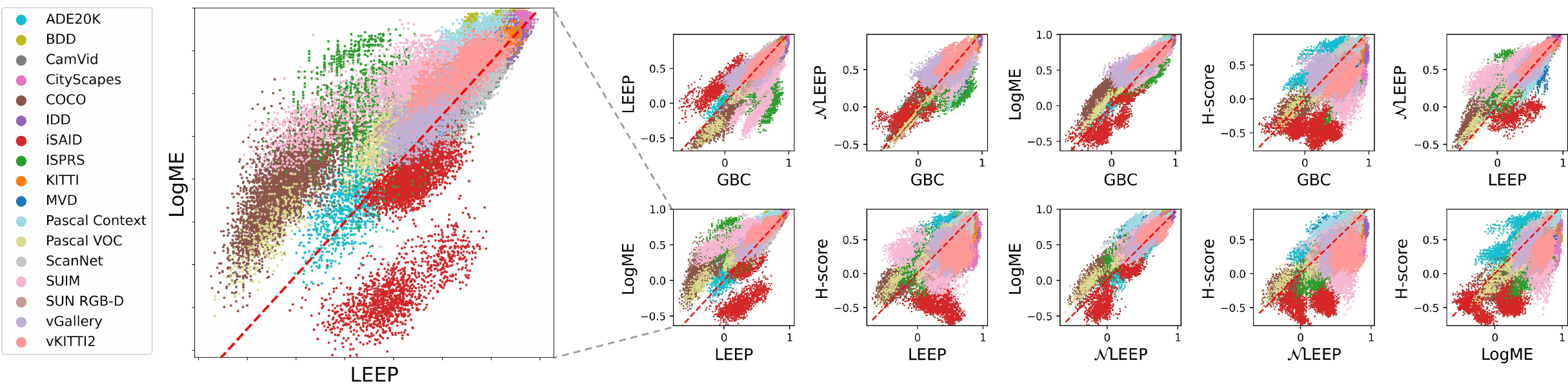} %
    \caption{
    Each point reports the quality of two transferability metrics in terms of $\tau_w$ (within a single experiment). In each plot we vary the source pool $\mathcal{S}$ and target dataset $T$. Points on the red line mean that the two transferability metrics have equal quality. For some comparisons the setup is not stable w.r.t $T$ and $\mathcal{S}$, as different points fall above and below the line, hence they produce divergent outcomes.}
    \label{fig:source-datasets-scatterplots-no-exp-avg}
\end{figure}

\para{Qualitative analysis.} \label{sec:semseg-qualitative-results}
We perform here the qualitative analysis described in Sec.\ref{sec:qualitative-method}. We fix the evaluation measure $E$ to be $\tau_w$ and visualize experiments with variations of source pool $\mathcal{S}$ and target dataset $T$. Fig. \ref{fig:source-datasets-scatterplots-no-exp-avg} shows comparisons for all possible pairs of transferability metrics.
If we compare \logme{} to \leep{} (Fig.\ref{fig:source-datasets-scatterplots-no-exp-avg}-left), the experiments evaluated on the target dataset $T= \textrm{iSAID}$ (red) form a group below the red line, indicating \leep{} consistently outperforms \logme{}.
Conversely, for both $T=\textrm{COCO}$ (brown) and $T=\textrm{SUIM}$ (pink), \logme{} consistently outperforms \leep{}. 
This shows the experimental instability w.r.t. variations of $T$.
If we look at $T=\textrm{ADE}$ (light blue), the experiments are centered on the red line and stray quite far from it on {\em both sides}. This means that contradicting outcomes are found also when varying the source pool $\mathcal{S}$.
The same observation holds for $T=\textrm{vGallery}$ (purple).

\begin{figure}[tp]
\setlength{\belowcaptionskip}{1mm} %
\begin{subfigure}[b]{\textwidth}
    \captionsetup{skip=0pt} %
    \centering
    \includegraphics[width=\textwidth]{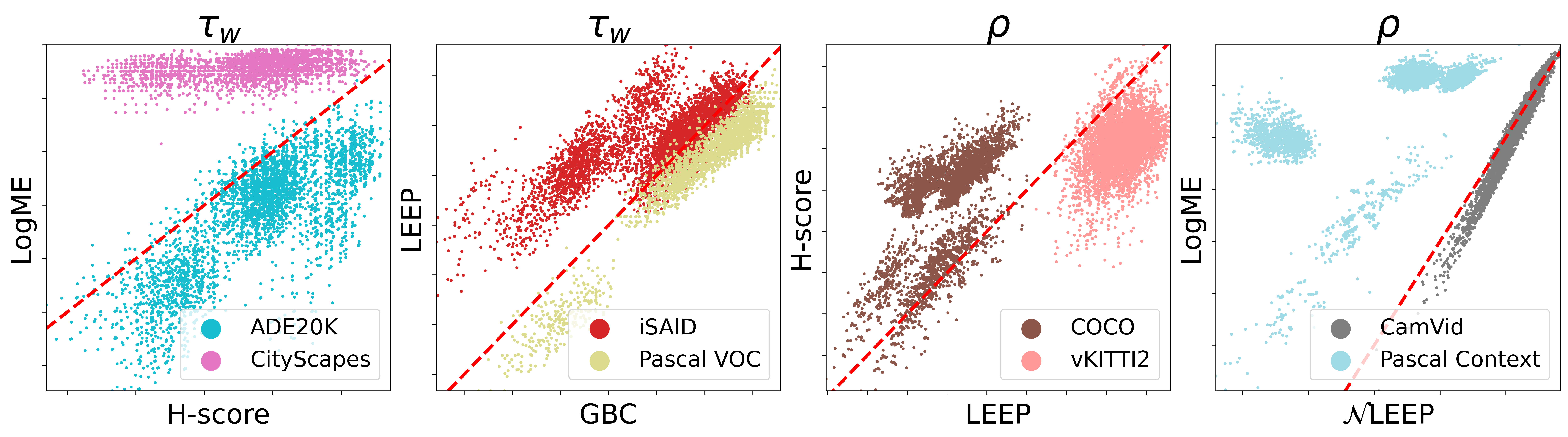}
    \caption{Effects of varying the target dataset $T$.
    }
    \label{fig:source_datasets_influence_targets_pairs}
\end{subfigure}
\begin{subfigure}[b]{\textwidth}
    \captionsetup{skip=0pt} %
    \centering
    \includegraphics[trim=5 130 5 90, clip, width=\textwidth]{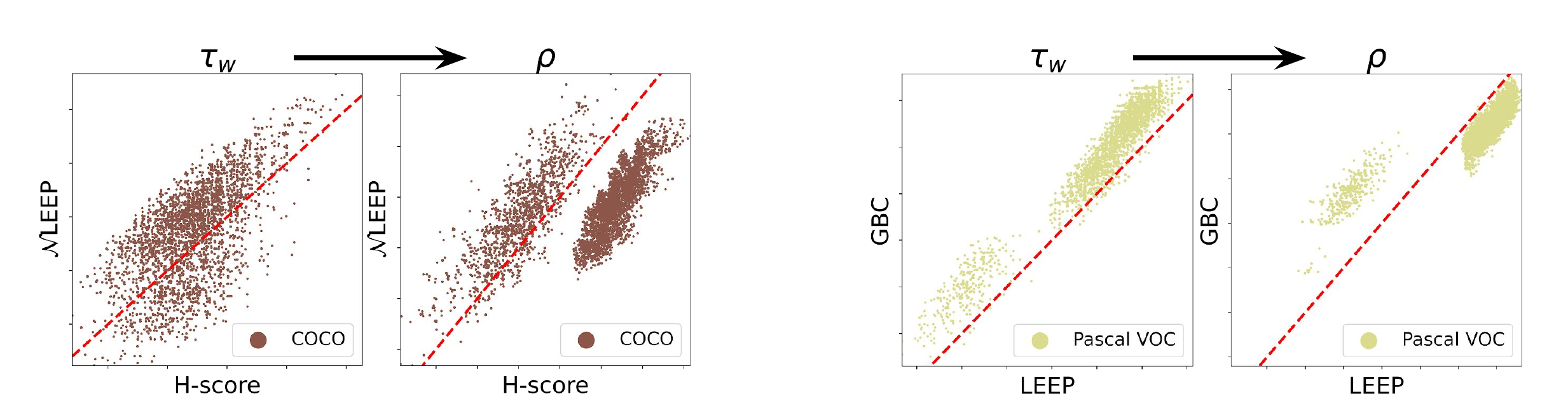}
    \caption{Effects of varying the evaluation measure $E$.}
    \label{fig:source_datasets_influence_evaluation_measures}
\end{subfigure}
\begin{subfigure}[b]{\textwidth}
    \captionsetup{skip=0pt} %
    \centering
    \includegraphics[width=\textwidth]{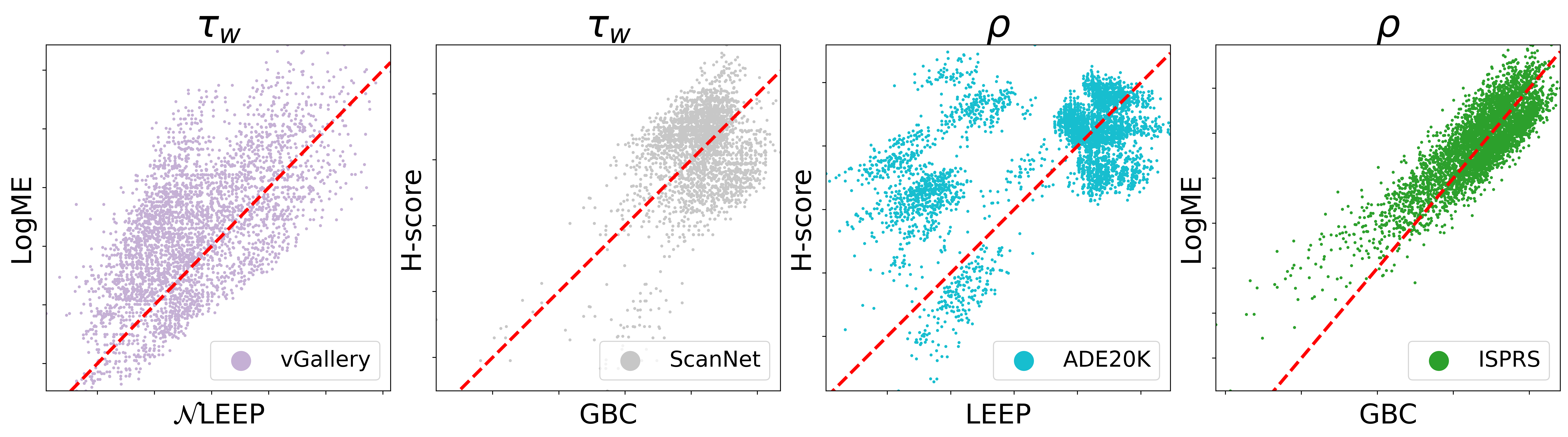}
    \caption{Effects of varying the source pool $\mathcal{S}$.}
    \label{fig:source_datasets_influence_source_models}
\end{subfigure}
\caption{Examples of how changing experimental setup components produces different winning metrics across experiments.}
\label{fig:source_datasets_influence_components}
\end{figure}

In fact, we can find many cases where the outcome of an experiment is wildly different even when varying only a single setup component (Fig. \ref{fig:source_datasets_influence_components}).
In Fig.~\ref{fig:source_datasets_influence_components}a, we see that for $T=\textrm{ADE20K}$ \hscore{} outperforms \logme{}, whereas for $T=\textrm{CityScapes}$ the opposite is true.
Next, Fig.~\ref{fig:source_datasets_influence_components}b show two examples where changing only the evaluation measure $E$ leads to conflicting outcomes.
On $T=\textrm{COCO}$, for $E=\tau_W$ \nleep{} is better than \hscore{} (most points are above the line), whereas for $E=\rho$ the opposite is true (large dense cluster below the line).
A similar contradicting result is observed on $T=\textrm{Pascal VOC}$ for \leep{} and \gbc{}.
Finally, we look at varying the source pool $\mathcal{S}$ in Fig.~\ref{fig:source_datasets_influence_components}c.
We see that in each of these example plots the points are scattered on both sides of the $x=y$ line, indicating contradicting outcomes within each plot.

To conclude, we observed that varying even just a single component of an experimental setup can produce conflicting results. This means it is important to consider variations of all components when designing experiments to evaluate transferability metrics.

\para{Quantitative analysis.} \label{sec:semseg-quantitative-results}
We now apply the \emph{Setup Stability} (SS) score defined in Sec. \ref{sec:quantitative-method}. Tab. \ref{tab:robustness-score} shows results when varying each component of the experimental setup, aggregating results over our experiments.
Variations of the target dataset $T$ are the main factor of instability: with a low Setup Stability (SS) score of 0.23, most experiments have different outcomes as to which transferability metric performs better.
Results are more stable to variations in the evaluation measure $E$ (0.53 SS), and most stable to variations in the source pool $\mathcal{S}$ (0.60 SS).

\begin{table}[ptb]
\begin{subtable}[b]{0.37\textwidth}
\centering
\resizebox{!}{12mm}{
\begin{tabular}{c|c}
  & \textbf{SS score} \\ \hline 
\textbf{$\neq$ T} & 0.23 \\
\textbf{$\neq$ E} & 0.53\\
 \textbf{$\neq \mathcal{S}$} & 0.60\\
\end{tabular}
}
\subcaption{}
\label{tab:robustness-score}
\end{subtable}
\hfill
\begin{subtable}[b]{0.62\textwidth}
\centering
\resizebox{!}{12mm}{
\begin{tabular}{c|c|c|c|c|c}
 & \textbf{\%W $\tau_w$} & \textbf{\%W $\tau$} & \textbf{\%W $p$} & \textbf{\%W $Rel@1$} & \textbf{Avg}\\ 
\hline
\textbf{\gbc{}}  & {20} & {16} & {10} & {5} & {13} \\
\textbf{\leep{}}  & {11} & {13} & {11} & {2} & {9} \\
\textbf{\nleep{}}  & {7} & {3} & {5} & {2} & {4} \\
\textbf{\logme{}}  & {\textbf{52}} & {\textbf{55}} & {\textbf{68}} & {\textbf{49}} & {\textbf{56}} \\
\textbf{\hscore{}}  & {10} & {13} & {6} & {42} & {18} \\
\end{tabular}
}
\subcaption{}
\label{tab:percentage-wins-dataset-selection}
\end{subtable}
\caption{Aggregate quantitative analysis for Sec.~\ref{sec:semseg}.
(a) \emph{Setup Stability} (SS) score for variations of source pool $\mathcal{S}$, target dataset $T$ or evaluation measure $E$.
(b) \emph{Win rate} (\textbf{\%W}) for each transferability metric and evaluation measure. \logme{} is on average the best metric.}
\label{tab:datasets-quantitative-analysis}
\end{table}

Finally, we compute the \emph{win rate} of each transferability metric (Sec.~\ref{sec:quantitative-method}) .
Results split by evaluation measure are shown in Tab.\ref{tab:percentage-wins-dataset-selection}.
\logme{} outperforms the other metrics on average 56\% of the times. In contrast, all other metrics have much lower winning rates (4-18\%).
Hence, by aggregating over all variations across setup components, we can reduce the experimental uncertainty and reveal that, on average, \logme{} is the best transferability metric for selecting source datasets.
In contrast, when evaluating on a limited set of experiments as in previous works (Tab.\ref{tab:setup_comparison}), results vary wildly and lead to contradictory conclusions (Fig. \ref{fig:source-datasets-scatterplots-no-exp-avg} and Fig. \ref{fig:source_datasets_influence_components}).

\section{Scenario 2: selecting a good source model architecture in image classification}
\label{sec:architecture-selection}

\para{Experimental setup.}
We consider the scenario of selecting good source model architectures to transfer from on the task of image classification, as in \cite{nguyen20icml,li21cvpr,you21icml,pandy2022transferability}.
We fix the source dataset to ImageNet. The goal is to select which architecture would lead to the best transfer learning results on a given target dataset.
We use $9$ target datasets: CIFAR10 \& 100~\cite{krizhevsky09}, Imagenette~\cite{imagenette}, Oxford IIIT Pets~\cite{parkhi12cvpr}, Caltech-USCD Birds 2011~\cite{CUB2011}, Stanford Dogs~\cite{khosla2011dogs}, Oxford Flowers 102~\cite{nilsback08ieee}, SUN-397~\cite{xiao10cvpr}, and DTD~\cite{cimpoi14cvpr}.
We consider a total of $19$ source architectures: ResNet-50 \& ResNet-101~\cite{he16cvpr}, ResNetV2-50, ResNetV2-101 \& ResNetV2-152~\cite{he2016identity}, DenseNet-121, DenseNet-169 \& DenseNet-201~\cite{huang2017densely}, MobileNet~\cite{howard2017mobilenets}, MobileNetV2~\cite{sandler18cvpr}, MobileNetV3~\cite{howard19iccv}, EfficientNetB0, EfficientNetB1, EfficientNetB2 \& EfficientNetB3~\cite{tan19icml}, NASNet Mobile~\cite{zoph18cvpr}, VGG16 \& VGG19~\cite{simonyan15iclr}, Xception~\cite{chollet17cvpr}.
We train 19 source models, each with a \emph{different} model architecture but all trained on the \emph{same} source dataset ImageNet~\cite{russakovsky15ijcv}.

\para{Setup variations.}
Now we create many setup variations.
To construct many different source pools, we consider all combinations of 14 out of 19 source models, resulting in $\binom{19}{14}$ source pools.
Finally, we consider four evaluation measures $E$, and the $9$ target datasets above.
Hence we have a total of $9 \times \binom{19}{14} \times 4 = 418$k experiments.

\begin{figure}[tp]
    \centering
    \includegraphics[trim=20 120 10 110, clip, width=\textwidth]{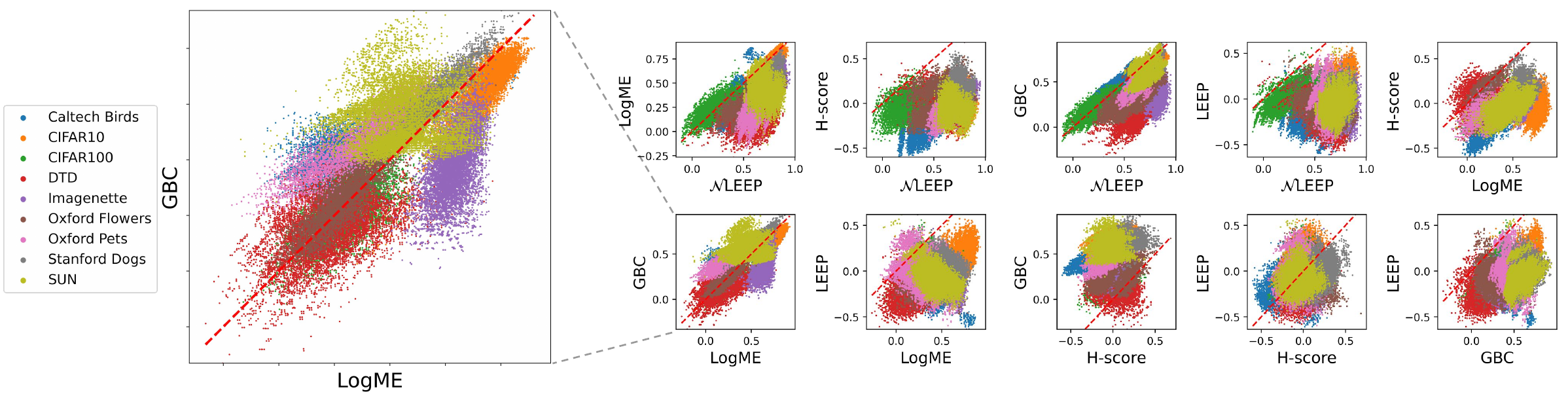}
    \caption{Each point reports the quality of two transferability metrics in terms of $\rho$ (within a single experiment).
    In each plot we vary the source pool $\mathcal{S}$ and target dataset $T$. Points on the red line mean that the two transferability metrics have equal quality. For some comparisons the setup is not stable w.r.t $T$ and $\mathcal{S}$, as different points fall above and below the line, hence they produce divergent outcomes (i.e \leep{} vs \hscore{} and \gbc{} vs \logme{}).}
    \label{fig:source-architectures-scatter-pairs}
\end{figure}

\para{Qualitative analysis.} \label{sec:archi-qualitative-results}
We perform the qualitative analysis described in Sec. \ref{sec:qualitative-method}. We fix the evaluation measure $E$ to be $\rho$ and we visualize experiments with variations of source pool $\mathcal{S}$ and target dataset $T$.
Fig. \ref{fig:source-architectures-scatter-pairs} shows comparisons for all possible pairs of transferability metrics.
Similar to Sec.~\ref{sec:semseg} we observe that many experiments within the same plot fall on either sides of the red line with a large spread. Hence outcomes are not stable. In particular, on the left-side plot, for $T=\textrm{Oxford Pets}$ (pink) \gbc{} consistently outperforms \logme{}, whereas for $T=\textrm{Imagenette}$ (purple) \logme{} consistently outperforms \gbc{}.

\begin{figure}[tp]
\begin{subfigure}[b]{\textwidth}
    \captionsetup{skip=0pt} %
    \centering
    \includegraphics[width=\textwidth]{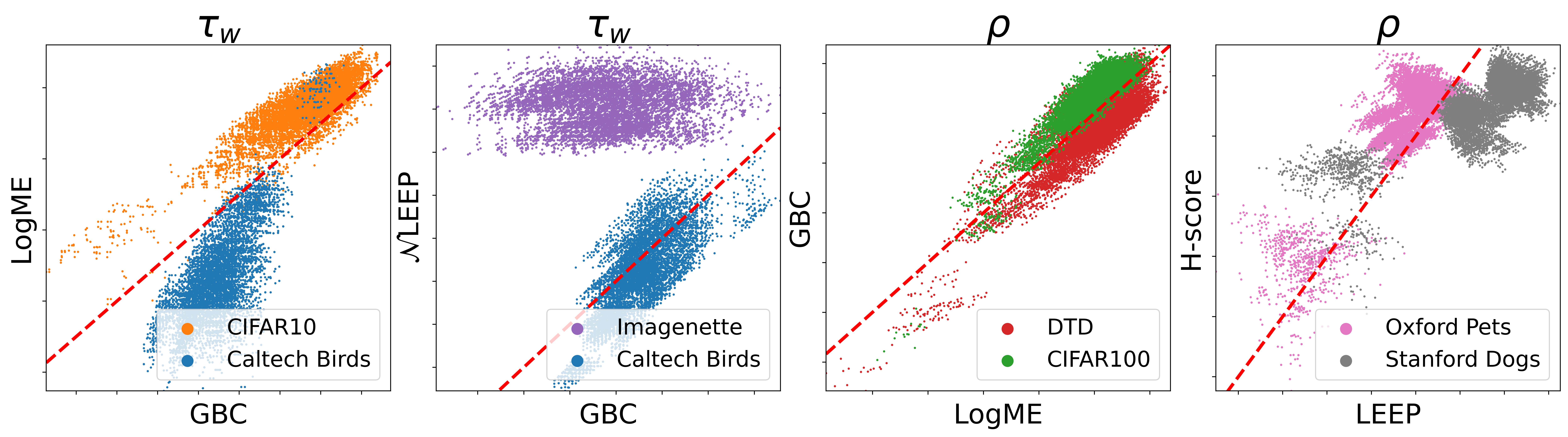}
    \caption{Effects of varying the target dataset $T$.}
    \label{fig:source_arch_influence_targets_pairs}
\end{subfigure}
\begin{subfigure}[b]{\textwidth}
    \captionsetup{skip=0pt} %
    \centering
    \includegraphics[trim=5 125 5 90, clip, width=\textwidth]{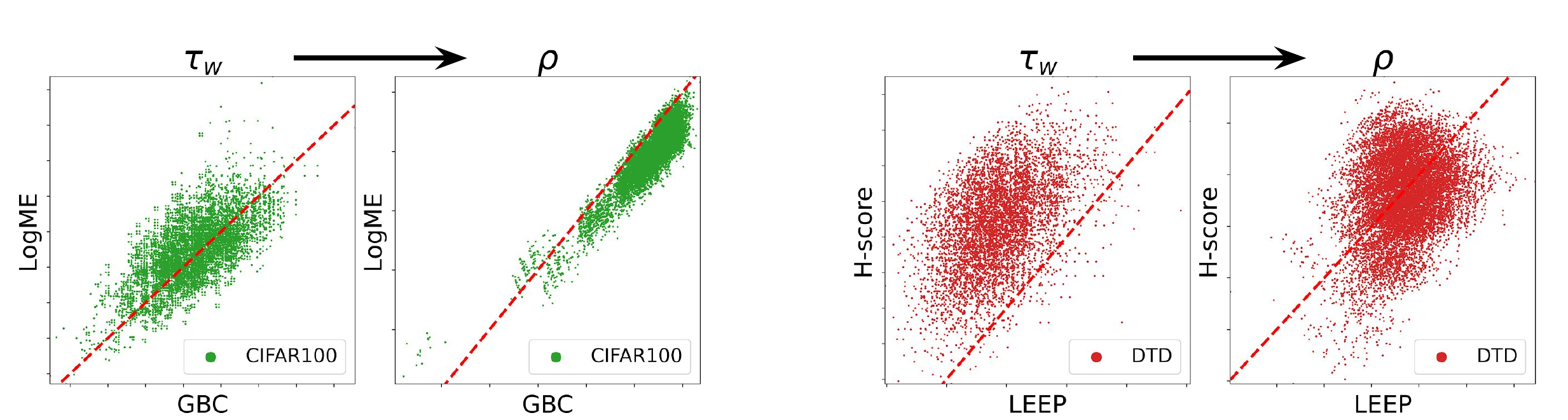}    
    \caption{Effects of varying the evaluation measure $E$.}
    \label{fig:source_arch_influence_evaluation_measures}
\end{subfigure}
\begin{subfigure}[b]{\textwidth}
    \captionsetup{skip=0pt} %
    \centering
    \includegraphics[width=\textwidth]{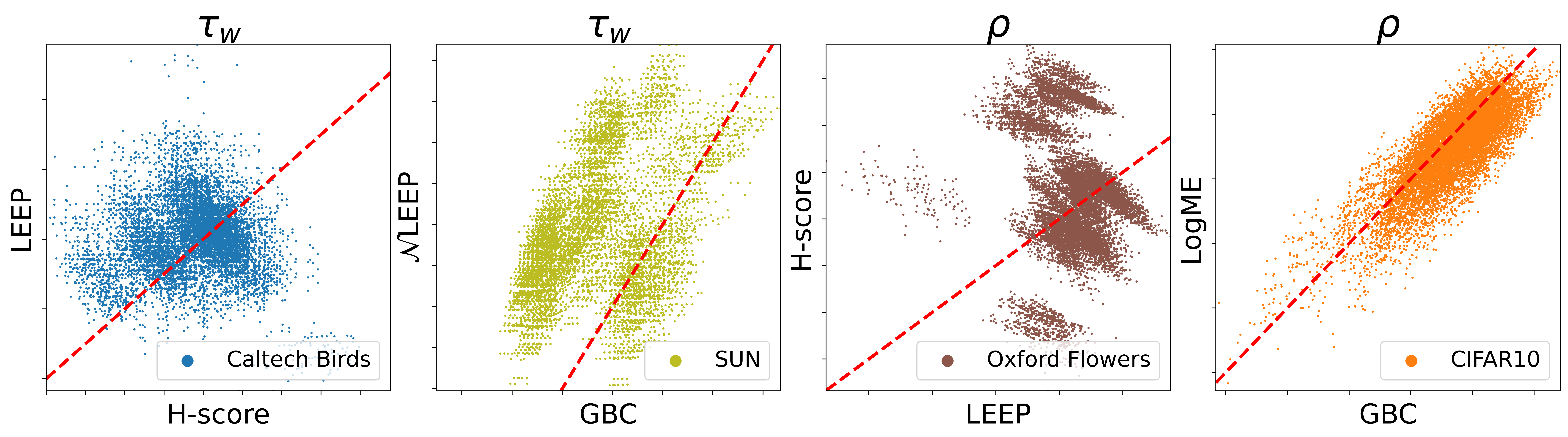}    
    \caption{Effects of varying the source pool $\mathcal{S}$.}
    \label{fig:source_arch_influence_source_models}
\end{subfigure}
\caption{Examples of how changing a single experimental setup component produces different winning metrics across experiments.}
\label{fig:source_architectures_influence_components}
\end{figure}

As before, there exists multiple examples where changing a single experimental setup component is sufficient to produce divergent outcomes across experiments (Fig. \ref{fig:source_architectures_influence_components}).
For example, in Fig.~\ref{fig:source_arch_influence_targets_pairs} for $T=\textrm{CIFAR10}$, \logme{} outperforms \gbc{}, while for $T=\textrm{Caltech Birds}$ the opposite is true.
Moreover, in Fig.~\ref{fig:source_arch_influence_evaluation_measures} we can clearly see the mass of the points move when changing $E$. 
For example, on the left \logme{} performs about equally to \gbc{} when using $E=\tau_w$, whereas \gbc{} is the best when using $E=\rho$.
Finally, when varying only the source pool $\mathcal{S}$ in 
Fig.~\ref{fig:source_arch_influence_source_models} we see points scattered on both sides of the line within each individual plot.
Hence the choice of each experimental setup component can have a large effect on the outcome.

\para{Quantitative analysis.}
\label{sec:archi-quantitative-results}
We now quantify the influence of varying experimental setup components by computing the \emph{Setup Stability} (SS) score defined in Sec.~\ref{sec:quantitative-method}. Results are shown in Tab.~\ref{tab:robustness-score-architectures}.
Varying the target dataset has the biggest impact (0.62 SS) on the instability of the experimental outcomes, followed by the evaluation measure (0.69 SS) and the source pool (0.80 SS).
This is in line with the observations in Sec.~\ref{sec:semseg}, except that the overall scores are  higher, indicating that the outcomes across experiments are broadly more consistent. %

Tab.\ref{tab:percentage-wins-architecture-selection} shows the \emph{win rate} of each transferability metric against all others, for each evaluation measure.
\nleep{} outperforms on average the other metrics in 73\% of the experiments,
while \leep{} and \hscore{} performs poorly (1 and 3\% winning rates, respectively).
Hence, by aggregating over many experiments we reveal a clear winner transferability metric in this scenario: \nleep{}. Moreover, \logme{} can be considered a good second place since it is the best for $E=\textit{Rel}@1$.

\begin{table}[tbp]
\begin{subtable}[b]{0.37\textwidth}
\centering
\resizebox{!}{12mm}{
\begin{tabular}{c|c}
  & \textbf{SS score} \\ \hline 
\textbf{$\neq$ T} & 0.62 \\
\textbf{$\neq$ E} & 0.69\\
 \textbf{$\neq \mathcal{S}$} & 0.80\\
\end{tabular}
}
\subcaption{}
\label{tab:robustness-score-architectures}
\end{subtable}
\hfill
\begin{subtable}[b]{0.62\textwidth}
\centering
\resizebox{!}{12mm}{
\begin{tabular}{c|c|c|c|c|c}
 & \textbf{\%W $\tau_w$} & \textbf{\%W $\tau$} & \textbf{\%W $p$} & \textbf{\%W $Rel@1$} & \textbf{Avg}\\ 
\hline
\textbf{\gbc{}}  & {12} & {4} & {5} & {4} & {6} \\
\textbf{\leep{}}  & {0} & {0} & {0} & {2} & {1} \\
\textbf{\nleep{}} & {\textbf{80}} & {\textbf{91}} & {\textbf{93}} & {30} & {\textbf{73}} \\
\textbf{\logme{}}  & {8} & {5} & {2} & {\textbf{53}} & {17} \\
\textbf{\hscore{}}  & {0} & {0} & {0} & {11} & {3} \\
\end{tabular}
}
\subcaption{}
\label{tab:percentage-wins-architecture-selection}
\end{subtable}
\caption{
Aggregate quantitative analysis for Sec.~\ref{sec:architecture-selection}.
(a) \emph{Setup Stability} (SS) score for variations of $\mathcal{S}$, $T$ \& $E$.
(b) \emph{Win rate} (\textbf{\%W}) for each transferability metric and evaluation measure. \nleep{} is on average the best metric.
}
\label{tab:archy-quantitative-analysis}
\end{table}

Importantly, the conclusion in this scenario differs substantially from the previous scenario: In Sec.~\ref{sec:semseg}, \logme{} was the best while \nleep{} was the worst. Now, \nleep{} is the best while \logme{} is a good second. Hence even though we aggregate over many experiments, we still find different conclusions for even larger changes in our setup ({\em scenarios}): changing both the nature of the source pool (i.e. different source datasets vs different source model architectures) and the nature of the task (semantic segmentation vs image classification).

\newpage
\section{Bonus Scenario}
\label{sec:target-selection}

As a somewhat separate investigation, we take a closer look at a commonly used scenario where the source model is fixed and many target tasks are constructed by sub-sampling classes from a single large target dataset~\cite{pandy2022transferability,nguyen20icml,tan21cvpr,tan21cvpr}. 
This scenario is interesting because of the curious results reported in previous works, including:
(1) some transferability metrics, including \leep{} and \nleep{}, have a bias towards the number of target classes~\cite{ibrahim2021newer};
(2) the performance of \logme{} is unstable: it delivers sometimes the best $\tau_w$ and sometimes the worst, even negative, $\tau_w$~\cite{pandy2022transferability}.
This might lead to misleading results, and hence we investigate here the experimental stability of this scenario.
In particular, we compare two different target dataset sub-sampling procedures and investigate the transferability metrics bias towards the number of target classes sampled.

An experiment in this scenario consists of three components: 
(1) A source model $S$.
(2) A target dataset, out of which we construct a \emph{target dataset pool} $\mathcal{T}$.  
(3) An evaluation measure $E$.

\textbf{Experimental setup and constructing $\mathcal{T}$.}
It is common practice to construct $\mathcal{T}$ by sampling uniformly between $2\%$ and a $100\%$ of the target classes~\cite{pandy2022transferability,nguyen20icml}. This results in target datasets which vary in the number of classes.
We compare this approach to uniformly sampling $50\%$ of the target classes, resulting in target datasets with an \emph{equal} number of classes.
In both cases we use all images for the selected classes for training and testing. 

We use as source model $S$ a ResNet50 pre-trained on ImageNet. 
We consider four target datasets: CIFAR100~\cite{krizhevsky09}, Stanford Dogs~\cite{khosla2011dogs}, Sun397~\cite{xiao10cvpr}, and Oxford Flowers 102~\cite{nilsback08ieee}.
For each target dataset we construct a pool $\mathcal{T}$ of $100$ datasets. The transferability metrics are evaluated as described in Sec.~\ref{sec:background}.

\textbf{Baseline transferability metric: NumC.}
For the purpose of our investigation, we now define a new baseline transferability metric.
Intuitively, target datasets that contain more classes are more complex than target datasets with fewer classes. 
Therefore, to determine to what extent is transferability is \emph{trivially} explained by the number of classes in a target dataset, we define the \emph{NumC} transferability metric as simply returning their number.

\textbf{Results.} 
The results presented in Tab.~\ref{tab:tar-rand} show that \textit{NumC} performs on par with the best transferability metrics (LEEP, GBC, \nleep{}) on all datasets in terms of $\tau_w$, and outperforms two other metrics (LogMe, H-score) on average. This highlights how this experimental setting rewards a bias towards the number of target classes.
We also note that while LogMe is the best performing metric in Sec.~\ref{sec:semseg-quantitative-results}, 
it now is the worst and has even negative $\tau_w$ in some cases (Tab.~\ref{tab:tar-rand}). 
In contrast, if we fix the number of target classes across all datasets in the pool, suddenly \logme{} delivers decent performance and beats \leep{} and \hscore{}. Of course, in this class-equalized setting, the trivial method \emph{NumC} becomes unusable.

To conclude, our findings suggest that a scenario where the target dataset pool $\mathcal{T}$ is constructed by sampling a variable number of classes is not suitable for evaluating transferability metrics. 
Instead, it is preferable to sample a fixed number of classes across the whole target pool $\mathcal{T}$.

If we look at the overall winning transferability metric, we find that \gbc{} works best in this scenario. 
Interestingly, this is again different from the winning metric in Sec.~\ref{sec:semseg} (\logme{}) and Sec.~\ref{sec:architecture-selection} (\nleep{}).

\begin{table}[t]
\centering
    \begin{subtable}[b]{0.75\textwidth}
    \resizebox{\textwidth}{!}{
        \begin{tabular}{c|cccccc}
        \hline
                          & \textbf{\gbc{}} & \textbf{\leep{}} & \textbf{\nleep{}} & \textbf{\logme{}} & \textbf{\hscore{}} & \textbf{NumC} \\
        \textbf{CIFAR100} & \textbf{0.935}        & 0.909         & 0.933          & -0.787        & 0.686            &   0.899 \\
        \textbf{Dogs} & 0.948        & \textbf{0.949}         & 0.936          & -0.789         & 0.200            &   0.913  \\
        \textbf{SUN} & \textbf{0.960}        & 0.947        & 0.950          & -0.920         & 0.473            &   0.953 \\
        \textbf{Flowers} & \textbf{0.748}        & 0.699         & 0.712         & -0.628         & -0.642           &   0.697\\ \hline
        \textbf{Average} &  \textbf{0.898}        & 0.876         & 0.883        & -0.781         & 0.179           &   0.866\\ \hline
        \end{tabular}
    }
    \caption{\footnotesize Sampling between $2\%$ and $100\%$ classes}
    \label{tab:tar-rand}
    \end{subtable}
    \begin{subtable}[b]{0.75\textwidth}
    \resizebox{\textwidth}{!}{
        \begin{tabular}{c|cccccc}
        \hline
                          & \textbf{\gbc{}} & \textbf{\leep{}} & \textbf{\nleep{}} & \textbf{\logme{}} & \textbf{\hscore{}} & \textbf{NumC} \\
        \textbf{CIFAR100} & 0.580       & 0.304         & 0.470          & \textbf{0.675}        & 0.386            &   - \\
        \textbf{Dogs} & \textbf{0.740}        & 0.703         & 0.611          & 0.521         & 0.191           &   -  \\
        \textbf{SUN} & 0.509        & 0.421         & \textbf{0.681}          & 0.263         & 0.256            &   - \\
        \textbf{Flowers} & 0.344        & 0.266         & 0.341       & \textbf{0.388}         & 0.167           &   -\\ \hline
        \textbf{Average} & \textbf{0.543}        & 0.424        & 0.526         & 0.462         & 0.250          &   -\\ \hline
        \end{tabular}
    }
    \caption{\footnotesize Sampling always $50\%$ classes}
    \label{tab:tar-fixed}
    \end{subtable}    
    \vspace{0.3cm}
\caption{$\tau_w$ performance of transferability metrics in ranking 100 randomly subsampled datasets out of a single large target dataset. We compare two different sampling strategies: uniformly sampling $2\!-\!100\%$ of the target classes (\ref{tab:tar-rand}) or always uniformly sampling $50\%$ of the target classes (\ref{tab:tar-fixed}). We repeat the experiment for each of 4 target datasets in turn (rows)}
\label{tab:target_selection}
\end{table}

\section{Conclusion}
Our work investigates for the first time the stability of experiments to evaluate transferability metrics. We base our investigation on a systematic analysis of over 715k experiments.
We show that even small variations across experiment setups leads to divergent conclusions about the superiority of a transferability metric over another. Then we improve stability by aggregating outcomes across many experiments.
As a result, we reveal the superiority of \logme{} at selecting good source datasets to transfer from, and \nleep{} at selecting good source architectures.
Finally, in a third scenario about determining which target task benefits most from a given source model, we reveal that sampling a variable number of target classes is not suitable for evaluating transferability metrics, and that \gbc{} performs best.
Hence, no single transferability metric works best in all scenarios.

In this work all scenarios feature source and target tasks of identical types (\emph{i.e.}, both segmentation, or both classification).
However, the proposed experimental protocols and analysis methods are not restricted to this setting. Hence, we leave to future work investigations of the experimental stability of transferability metrics when source and target task types differ.

\bibliographystyle{splncs04}
\bibliography{shortstrings,loco,loco_extra}
\end{document}